\RequirePackage{tikz}

\documentclass[pdflatex,sn-mathphys]{sn-jnl}

\jyear{2023}%

\theoremstyle{thmstyleone}%
\theoremstyle{thmstyletwo}%

\theoremstyle{thmstylethree}%

\definecolor{tml}{rgb}{0.7,0.9,0.7}

\definecolor{kostas}{rgb}{0.9,0.7,0.6}

\usepackage{graphicx}
\usepackage{subcaption}
\usepackage{amsmath}
\usepackage{xcolor}
\usepackage{color, soul}
\usepackage{tabularx}
\usepackage{adjustbox}
\usepackage{floatflt}

\usepackage{tikz}
\usepackage{pgfplots}
\usetikzlibrary{backgrounds}

\makeatletter
\newif\if@showgrid@grid
\newif\if@showgrid@left
\newif\if@showgrid@right
\newif\if@showgrid@below
\newif\if@showgrid@above
\tikzset{%
    every show grid/.style={},
    show grid/.style={execute at end picture={\@showgrid{grid=true,#1}}},%
    show grid/.default={true},
    show grid/.cd,
    labels/.style={font={\sffamily\small},help lines},
    xlabels/.style={},
    ylabels/.style={},
    keep bb/.code={\useasboundingbox (current bounding box.south west) rectangle (current bounding box.north west);},
    true/.style={left,below},
    false/.style={left=false,right=false,above=false,below=false,grid=false},
    none/.style={left=false,right=false,above=false,below=false},
    all/.style={left=true,right=true,above=true,below=true},
    grid/.is if=@showgrid@grid,
    left/.is if=@showgrid@left,
    right/.is if=@showgrid@right,
    below/.is if=@showgrid@below,
    above/.is if=@showgrid@above,
    false,
}

\def\@showgrid#1{%
    \begin{scope}[every show grid,show grid/.cd,#1]
    \if@showgrid@grid
    \begin{pgfonlayer}{background}
    \draw [help lines]
        (current bounding box.south west) grid
        (current bounding box.north east);
    \pgfpointxy{1}{1}%
    \edef\xs{\the\pgf@x}%
    \edef\ys{\the\pgf@y}%
    \pgfpointanchor{current bounding box}{south west}
    \edef\xa{\the\pgf@x}%
    \edef\ya{\the\pgf@y}%
    \pgfpointanchor{current bounding box}{north east}
    \edef\xb{\the\pgf@x}%
    \edef\yb{\the\pgf@y}%
    \pgfmathtruncatemacro\xbeg{ceil(\xa/\xs)}
    \pgfmathtruncatemacro\xend{floor(\xb/\xs)}
    \if@showgrid@below
    \foreach \X in {\xbeg,...,\xend} {
        \node [below,show grid/labels,show grid/xlabels] at (\X,\ya) {\X};
    }
    \fi
    \if@showgrid@above
    \foreach \X in {\xbeg,...,\xend} {
        \node [above,show grid/labels,show grid/xlabels] at (\X,\yb) {\X};
    }
    \fi
    \pgfmathtruncatemacro\ybeg{ceil(\ya/\ys)}
    \pgfmathtruncatemacro\yend{floor(\yb/\ys)}
    \if@showgrid@left
    \foreach \Y in {\ybeg,...,\yend} {
        \node [left,show grid/labels,show grid/ylabels] at (\xa,\Y) {\Y};
    }
    \fi
    \if@showgrid@right
    \foreach \Y in {\ybeg,...,\yend} {
        \node [right,show grid/labels,show grid/ylabels] at (\xb,\Y) {\Y};
    }
    \fi
    \end{pgfonlayer}
    \fi
    \end{scope}
}
\makeatother
\tikzset{every show grid/.style={show grid/keep bb}}

\newcommand{\aes}[1]{\textcolor{black}{#1}}

\begin{document}

\title[Differentiable Methods in Visual Computing]{Differentiable Visual Computing for Inverse Problems and Machine Learning}


\author*[1,3]{\fnm{Andrew} \sur{Spielberg}}\email{aespielberg@seas.harvard.edu}

\author[2]{\fnm{Fangcheng} \sur{Zhong}}\email{fangcheng.zhong@cst.cam.ac.uk}

\author[3]{\fnm{Konstantinos} \sur{Rematas}}\email{krematas@google.com}

\author[4]{\fnm{Krishna Murthy} \sur{Jatavallabhula}}\email{jkrishna@mit.edu}

\author[2,3]{\fnm{Cengiz} \sur{Oztireli}}\email{aco41@cam.ac.uk}

\author[5]{\fnm{Tzu-Mao} \sur{Li}}\email{tzli@ucsd.edu
}

\author[6]{\fnm{Derek} \sur{Nowrouzezahrai}}\email{derek@cim.mcgill.ca}

\affil*[1]{\orgdiv{School of Engineering and Applied Sciences}, \orgname{Harvard University}, \orgaddress{\street{Massachusetts Hall}, \city{Cambridge}, \postcode{02138}, \state{MA}, \country{United States of America}}}

\affil[2]{\orgdiv{Computer Science and Technology}, \orgname{University of Cambridge}, \orgaddress{\street{The Old Schools, Trinity Ln, \city{Cambridge}, \postcode{CB2 1TN}, \country{United Kingdom}}}}

\affil[3]{\orgname{Google}, \orgaddress{\street{1600 Amphitheatre Parkway}, \city{Mountain View}, \postcode{94043}, \state{CA}, \country{United States of America}}}

\affil[4]{\orgdiv{Computer Science and Artificial Intelligence Laboratory}, \orgname{Massachusetts Institute of Technology}, \orgaddress{77 Massachusetts Ave}, \city{Cambridge}, \postcode{02139}, \state{MA}, \country{United States of America}}

\affil[5]{\orgdiv{Computer Science \& Engineering}, \orgname{University of California --- San Diego}, \orgaddress{\street{5998 Alcala Park Way}, \city{San Diego}, \postcode{92110}, \state{CA}, \country{United States of America}}}

\affil[6]{\orgdiv{Electrical \& Computer Engineering}, \orgname{McGill University}, \orgaddress{\street{845 Rue Sherbrooke O}, \city{Montr\'eal}, \postcode{H3A 0G4}, \state{QC}, \country{Canada}}}


\abstract{

Modern 3D computer graphics technologies are able to reproduce the dynamics and appearance of real world environments and phenomena, building atop theoretical models in applied mathematics, statistics, and physics.  These methods are applied in architectural design and visualization, biological imaging, and visual effects. Differentiable methods, instead, aim to determine how graphics outputs (i.e., the real world dynamics or appearance) change when the environment changes.  We survey this growing body of work and propose a holistic and unified \aes{differentiable visual computing} pipeline. Differentiable visual computing can \aes{be leveraged to efficiently} solve otherwise intractable problems in physical inference, optimal control, object detection and scene understanding, computational design, manufacturing, autonomous vehicles, and robotics. Any application that can benefit from an understanding of the underlying dynamics of the real world stands to benefit significantly from a differentiable graphics treatment.

We draw parallels between the well-established computer graphics pipeline and a unified differentiable graphics pipeline, targeting consumers, practitioners and researchers. The breadth of fields that these pipelines draws upon --- and are of interest to --- includes the physical sciences, data sciences, vision and graphics, machine learning, and adjacent mathematical and computing communities.

}




\maketitle

\section{Introduction}

Originally designed for applications in computer graphics, visual computing (VC) methods synthesize information about physical and virtual worlds, using prescribed algorithms optimized for spatial computing.  VC is used to analyze geometry, physically simulate solids, fluids, and other media, and render the world \textit{via} optical techniques.  These fine-tuned computations that operate explicitly on a given input solve so-called forward problems, VC excels at.  By contrast, deep learning (DL) allows for  the construction of general algorithmic models, side stepping the need for a purely first principles-based approach to problem solving. DL is powered by highly parameterized neural network architectures --- universal function approximators --- and gradient-based search algorithms which can efficiently search that large parameter space for optimal models.  This approach is predicated by neural network differentiability, the requirement that analytic derivatives of a given problem's task metric can be computed with respect to neural network's parameters.  Neural networks excel when an explicit model is not known, and neural network training solves an inverse problem in which a model is computed from data.

While VC provides a strong inductive bias about the dynamics of real-world phenomena --- one that would otherwise have to be learned from scratch in a pure DL context --- its inability to adapt its mathematical models based on observations of real-world phenomena precludes its direct integration into larger DL-based systems. Meanwhile, DL is adaptive but requires a sufficiently expressive neural network, high-quality and diverse data observations of the underlying dynamics, and time to train the model to formulate (potentially complex) relationships between observations and dynamics. Differentiable visual computing (DVC) is a nascent family of approaches that aims to bridge this divide.

DVC combines the accurate, compact and explicit models of reality with system parameterization and gradients needed to adapt their parameters, \textit{i.e.}, using gradient-based optimization.  DVC \textit{constrains} otherwise general, but data-inefficient DL methods to the boundaries of physical reality.  As such, the combination of DL- and DVC-based methods can significantly improve the data efficiency, speed and accuracy with which machine learning can be applied to inference problems in real-world physical systems. Despite its infancy, DVC is unlocking applications in areas such as physical reasoning from video, biomechanics modeling, robot control, automated design for manufacturing, and autonomous driving.

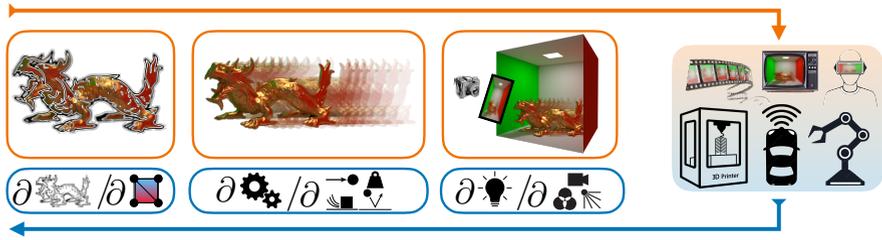
\begin{figure*}
	\begin{tikzpicture}
		\node[anchor=south west,inner sep=0] at (0,0) { 
		    	\includegraphics[width=\textwidth,page=1,trim=5.0cm 9cm 10cm 2.0cm, clip]{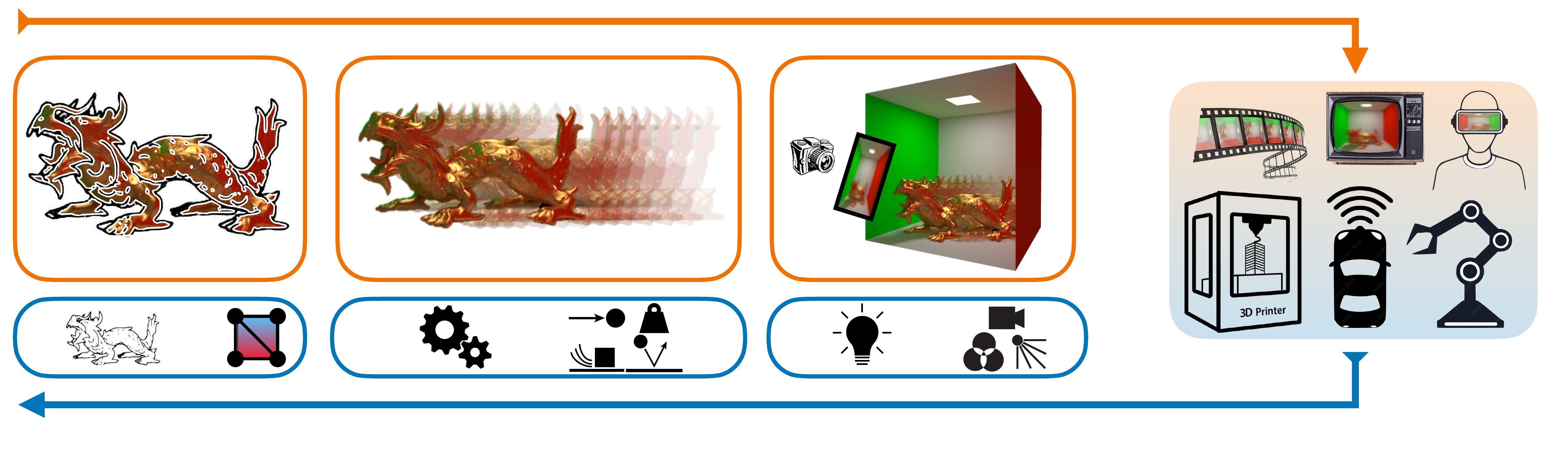} 
		}; 
		\node[scale=1.325,anchor=south west] at (0.015,0.695) { $\partial$};
		\node[scale=1.325,anchor=south west] at (1.125,0.6325) { $/$};
		\node[scale=1.325,anchor=south west] at (1.25,0.695) { $\partial$};
		\node[scale=1.325,anchor=south west] at (2.7,0.735) { $\partial$};
		\node[scale=1.325,anchor=south west] at (3.625,0.6125) { $/$};
		\node[scale=1.325,anchor=south west] at (3.805,0.675) { $\partial$};
		\node[scale=1.325,anchor=south west] at (5.85,0.715) { $\partial$};
		\node[scale=1.35,anchor=south west] at (6.65,0.6125) { $/$};
		\node[scale=1.325,anchor=south west] at (6.805,0.675) { $\partial$};
	\end{tikzpicture}
	\vspace{-0.5cm}
\caption{\label{fig:pipeline}Forward and backward visual computing pipeline diagram. \textbf{Top:} visual computing steps through (left to right) the geometric representation, dynamical evolution, and image synthesis of virtual 3D scenes, resulting in (far right) output image sequences \aes{(top row) or inferences for applications such as digital manufacturing, autonomous driving, and industrial robotics (bottom row).} This forward pipeline can be used, e.g., for applications in visual effects, computer aided design, virtual reality, and medical imaging, to name a few. \textbf{Bottom:} the differentiable visual computing pipeline aims to compute sensitivities of the associated forward pipeline components (geometry processing, animation, and rendering) with respect variations in their inputs. Most immediately, such a pipeline can be aimed at a diversity of inverse problems, where knowledge of the physical processes underlying can be imbued into task-based learning objectives.}
\end{figure*}

We present our vision of a complete differentiable visual computing pipeline, \aes{shown in Fig. \ref{fig:pipeline}}  that combines advances in \textit{differentiable} geometry, animation, and physics not only to generate visual content, but to efficiently understand and improve it using gradient-based information.  We provide a holistic perspective on the spectrum of differentiable visual computing problems, relating them to their analogues in the computer graphics domain while highlighting the additional expressive power they can leverage through their marriage with modern deep learning techniques.  
Perhaps most significantly, we discuss the potential of differentiable visual computing to significantly reduce the amount of data modern deep learning methods require 
whilst improving their generalization and interpretability.

We proceed by reviewing preliminaries on differentiable programming; we then expand on the vision of the DVC pipeline and review state-of-the art methods for its constituent parts, and conclude with the state of the field, including infrastructure, challenges, and opportunities.  Throughout, we draw parallels in challenges shared between fields and opportunities for cross-pollination of approaches with DL.

\section{Preliminaries \aes{on Differentiable Models, Autodifferentiation, and Visual versus\ Learned Models}}

\begin{figure*}
\centering
         \includegraphics[width=0.95\textwidth, page=2,trim=0.8cm 7cm 12.5cm 3.0cm, clip] {figs/Nature-DiffSim-fig2.pdf} 
         \includegraphics[width=0.95\textwidth, page=1]{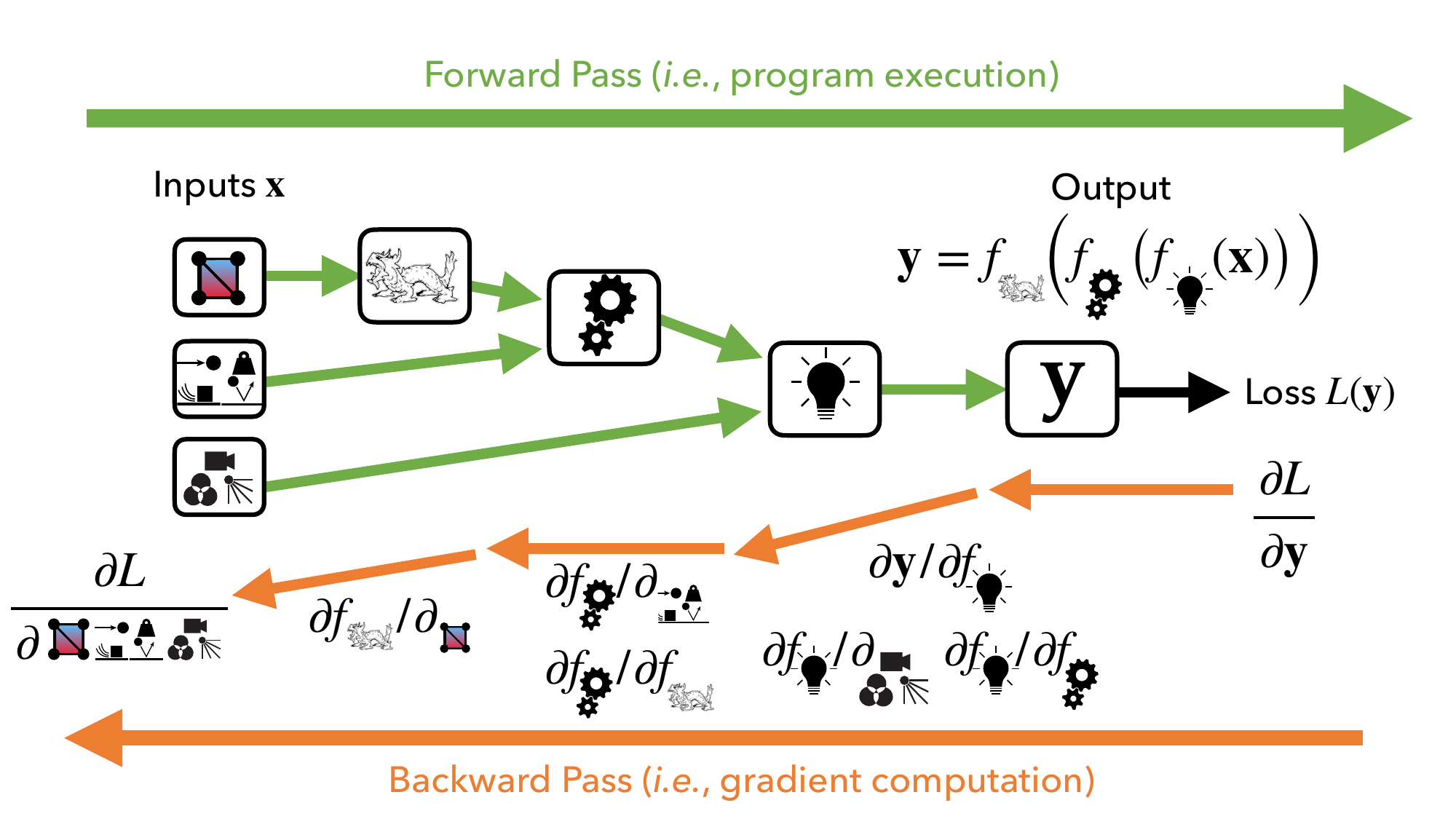}
\caption{Given a forward computational model -- i.e., expressed mathematically or programmatically -- such as the visual computing pipeline (\textbf{left}), we can compose the partial derivatives of the intermediate outputs of each stage of computation with respect to their inputs, in order to arrive at the total derivative of the final output with respect to the original inputs (\textbf{right}). Automatic differentiation \cite{griewank2008evaluating} and specialized methods can be used to transform forward models (e.g., such as the code block, \textbf{left}) into the backward model useful for gradient-based optimization. \aes{Currently prevalent competing approaches} replace the specialized visual computing processing blocks above with more standard non-linear computational blocks from neural network architectures, such as convolutional or fully-connected operations.}
\label{fig:backprop}
\end{figure*}

In the era of data abundance, the success of modern machine learning is predicated atop two key technologies: expressive differentiable computation models, and sophisticated gradient descent-based optimization routines that specialize these models to target applications/tasks.

Concretely, a model $f \colon \mathbf{x} \in \mathbb{R}^n \rightarrow \mathbf{y} \in \mathbb{R}^m$ parameterized by $\pmb{\theta} \in \mathbb{R}^p$ is said to be differentiable if derivatives of any $y \in \mathbf{y}$ can be evaluated with respect to any $x \in \mathbf{x}$ and $\theta \in \pmb{\theta}$. Note that $f$ could be an analytical function or a complex algorithm of indeterminate length (such as an optimization procedure).  See Fig. \ref{fig:backprop}. 
In practice, guarantees on gradient behavior -- \textit{e.g.}, that they be nonzero (in most places), sufficiently smooth, and efficient to compute --   are necessary for performant gradient-driven optimization.

Perhaps the most well-known differentiable models today are neural networks, employed in DL \aes{(for an extensive introduction to the topic, see \cite{goodfellow2016deep})}.  Typically constructed as high parameter compositions of differentiable primitives (layers), neural networks are employed -- with great success -- across a stunningly diverse set of tasks. Each network layer is designed such that the gradient of any output of the layer can be computed with respect to its inputs and its parameters.  Using the chain rule, gradients can then be computed across the entire network.  A typical workflow computes the gradient of a scalar output computation (often called a ``loss'') with respect to network parameters, conditioned on some input dataset.  The loss specifies a task through minimizing a penalty function (\emph{e.g.}, the error of a prediction $\mathbf{y}$ with respect to known ground truth $\hat{\mathbf{y}}$). Using these gradients,  optimization procedures can efficiently search a high-dimensional parameter space to minimize a loss and thus maximize task performance.

Since DL typically optimizes a single loss with respect to a very high-dimensional parameter set, backward-mode autodifferentiation (also called backpropagation) is employed.  Backpropagation computes the gradients of a loss with respect to its inputs by recursively computing and composing gradients backwards through the network layers, eventually generating gradients of the loss with respect to every network parameter.  See Fig. \ref{fig:backprop} for an illustration.  Computing the layer-wise Jacobian-vector products needed to compose gradients according to the chain rule requires knowing the value of the outputs of each layer.  Thus, this ``backward'' pass is preceded by a ``forward'' evaluation of the network (and loss), during which layer evaluations are computed and cached.  This procedure is memory intensive for deep and/or high-dimensional models, but keeps gradient computation runtime efficient.

The mathematical underpinnings of differentiable visual models mirror those of DL:  we can decompose a DVC model into a set of subcomputations (cf layers), whose relation is specified by a general directed acyclic computation graph (DAG, cf neural network). We can  similarly backpropagate through the visual model's DAG to compute gradients, assuming we first cache the outputs of each DAG node during forward evaluation steps. The building block computation nodes (cf neurons) in DVC models include geometry processing, physics-based animation, and image synthesis (more commonly referred to as rendering; see Fig.~\ref{fig:pipeline}.)  While most DVC models use backpropagation for differentiation, certain problems are served by algorithms that require gradients of many outputs of a model with respect to a single input/parameter, or the gradients of many outputs with respect to many inputs/parameters;
in these cases, forward and mixed differentiation schemes can be preferable.  We touch upon these alternate strategies throughout this survey.

The power of differentiable visual models is that they \aes{facilitate} efficient gradient computation \aes{which can significantly outperform finite differencing schemes in speed, accuracy, or both,} or the need to learn a differentiable model.  This saves time and diminishes the need for data, which can be expensive to gather or generate.  However, visual models are often more complex than neural networks; their gradients are more expensive to compute and their representations are even more nonlinear, which can lead to numerical issues \aes{and analytical search algorithms becoming trapped in local minima}.  
The remainder of this article takes a magnifying glass to how these visual models are constructed in order to increase their efficacy, as well as how they can be composed with other visual and DL models in cutting-edge real-world applications.

\section{\aes{The Forward and Backward Visual Computing Pipelines}}\label{pipeline}

The classic computer graphics pipeline proceeds by instantiating geometry, animating it as desired, and then rendering the geometry for visual appeal.  At a high level, this is not so different from the way scientists and engineers reason about Nature and design --- first defining a geometric representation for their problem, simulating its behavior, and then analyzing it based on (often visual) observation.  In fact, modern simulation-powered CAD software presents an example of such techniques operating simultaneously in the contexts of scientific analysis, design and engineering, and computer graphics.  In this section, we provide a high-level overview of the classical VC pipeline forward process and our vision of its differentiable analogue.

DVC enables each computer graphics process to be differentiable, as well as the end-to-end pipeline.  Imagine one wishes to design a scene for an animated fantasy film.  They might first specify the geometry of a dragon, animate its motion for a scene, and then render the dragon to be viewed on film.  Since each section (geometry, animation, rendering) is a differentiable process, one could optimize the dragon's shape, control, and lighting in order to make it look as bold, scary, bright, \textit{etc.} as possible (Fig. \ref{fig:pipeline}) by computing gradients of that objective and using them for numerical optimization. 
While end-to-end differentiable pipelines are not presently used in production, this vision is not a pipe dream, as differentiable methods are increasingly employed in domain-specific industrial tasks (\emph{e.g.} constraint satisfaction in CAD, optimal control in animation, accurate illumination in rendering).  

At the research level, however, differentiable methods across domains are already being combined for diverse applications. For example, 
\cite{ma2021diffaqua} co-optimized the design and behavior of simulated robotic fish, combining using differentiable geometry and animation.  
\cite{liu2019soft} combined differentiable rendering and geometry techniques to estimate shape from images.  
\cite{murthy2020gradsim} demonstrated inference of the materiality of physical objects directly from rendered video, while 
\cite{RempeContactDynamics2020} applied a similar approach to estimate human motion.  Recently, 
\cite{peng2021animatable} demonstrated the extraction of geometry, animation and rendering parameters from multiview video, effectively combining all elements of the differentiable pipeline.
Further, each of these works demonstrates how DVC layers can be seamlessly integrated with traditional DL models, which can be trained end-to-end or in their own standalone processes \cite{spielberg2019learning}.  The integration of DL models makes it possible to learn otherwise difficult-to-model phenomena (including difficult-to-model physics or user aesthetic preferences) and include these in optimization-driven computer graphics or science applications.

Unfortunately, the physical world is full of noise and discontinuities.  Modeling physical phenomena in a way that is accurate and useful (\emph{i.e.}, the \aes{correct} prior), but also differentiable, remains a core salient challenge. We next describe differentiable techniques in geometry, animation, and rendering, and describe the challenges and advances in each.

\section{Geometry}\label{geometry}

\begin{figure*}
	\begin{tikzpicture}
	    \node[anchor=south west,inner sep=0] at (0,0) {
        \includegraphics[width=\textwidth]{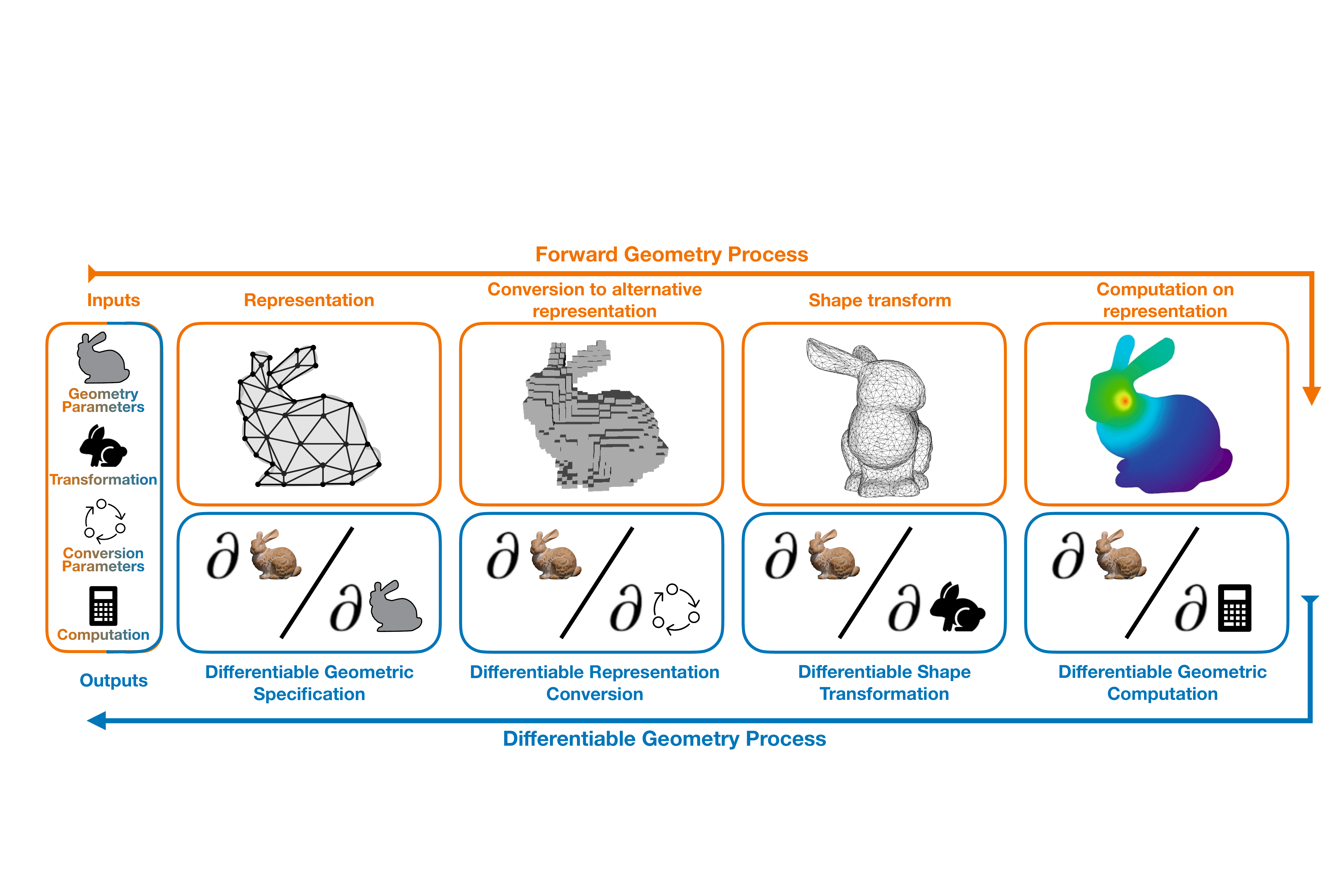}};
	\end{tikzpicture}
\caption{\label{fig:geometry}Forward and differentiable geometry processing. \textbf{Top:} geometry processing deals with problems surrounding (middle, left to right) the discrete representation, conversion between alternative representations, potential transformations, and computation of properties of the surfaces of geometric objects. \textbf{Bottom:} in the differentiable setting, gradients of (potentially continuous) geometric properties are sought with respect to (middle, right to left) measured signals on their manifolds, transformations applied to them, and the underlying discretization and representational parameterizations.}
\end{figure*}

\begin{figure*}
    \centering
    \includegraphics[width=\textwidth, page=2, clip, trim=0cm 5cm 8cm 0cm]{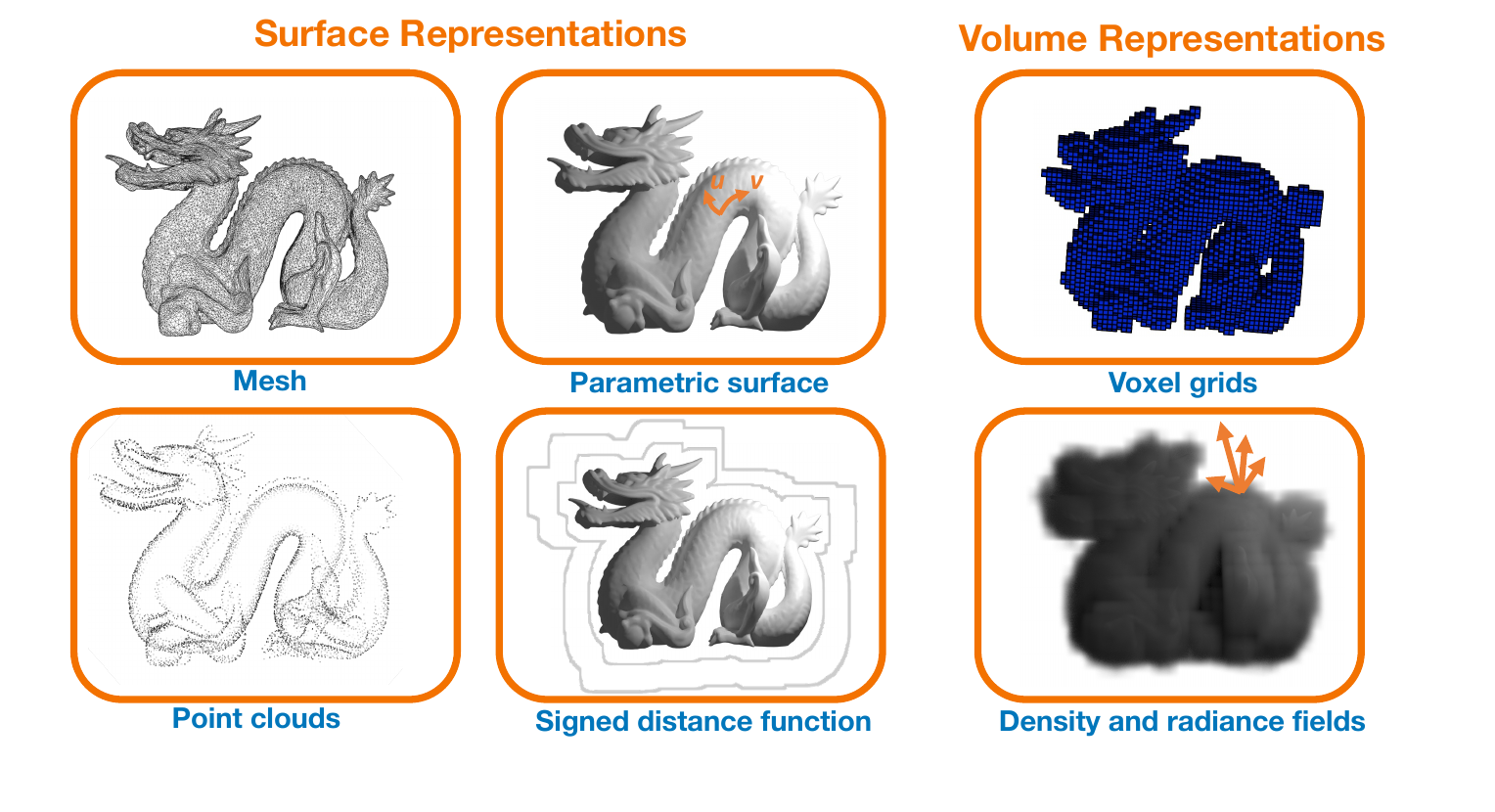}
    \caption{Visual illustration of common surface and volumetric geometric representations.}
    \label{fig:representation}
\end{figure*}

The VC pipeline begins with geometry.  For something to move or to be seen, it must exist, and geometry specifies where ``things'' exist and where they do not.  These ``things'' could be tangible, like an apple falling due to gravity, or matterless but still physical, like an area light source, or completely abstract\aes{, like a region for numerical integration.}  Regardless, the definition of a geometric object is always mathematical, with direct or algorithmic means of specification.

The ability to use gradients to optimize geometry is directly powerful in engineering applications such as surface reconstruction \cite{kazhdan2006poisson}, topology optimization \cite{sigmund2013topology}, designing printed objects with desired mechanical properties \cite{prevost2013make, bacher2014spin}, and robot morphology optimization \cite{ma2021diffaqua}.
Recent experimental CAD kernels have been introduced that allow for differentiating  through the geometry specification process \cite{banovic2018algorithmic, keeter2020massively, cascaval2021differentiable} and optimize for manufactured performance at design time.

Representations in geometry typically fall along three axes: 1) volumetric vs. surface, 2) explicit vs. implicit, 3) discrete vs. continuous, as illustrated in Fig. \ref{fig:representation}.
Volumetric representations specify an occupancy or density value for every position in a 3D volume. Such specification can be either discrete (\textit{e.g.} voxel grids~\cite{liu2020, yu2021plenoxels}) or continuous (e.g. occupancy/density field~\cite{Lombardi2019, mildenhall2020nerf}).
Volumetric representations are the most general form for characterizing a 3D scene and are robust in modeling complex scene geometry and topology such as hair, fabric, and smoke.  This generality comes at the expense of memory efficiency as each segment of space must be explicitly specified; further, volumetric representations lack certain precomputed information useful for applications, such as surface normals. 
Surface representations, on the other hand, explicitly or implicitly specify boundaries of in volumes.  These representations are compact and often more geometrically expressive compared to volumetric representations\aes{, but can impel additional computations, such as checking if a point is interior}.
Explicit representations can be unions of discrete boundary elements (\textit{e.g.} meshes~\cite{Nicolet2021})
or continuous functions (\textit{e.g.} parametric surfaces~\cite{groueix2018}, and splines~\cite{Cho2021})
while implicit approaches specify a surface by identifying the level set of a trivariate function (\textit{e.g.} a signed distance function).  \aes{While implicit representations are more general, explicit geometry allows for direct computation of a value (such as distance from a boundary) at any point in space.}
\aes{Meanwhile, discrete representations sacrifice precision for simplicity.}
Because each representation has its own pros and cons, they co-exist in VC\aes{; some hybrid methods operate across multiple representations \cite{chen2022tensorf, muller2022instant}.}

Representations have advantages and disadvantages in DVC as well, and geometric representations are further chosen according to the type of gradient information needed for the downstream task.  In practice, geometries can be (differentiably) converted between representations within an algorithm, and different representations can be used for analysis (training) and deployment (inference).
If a representation is or can be made continuous and smooth, the recipe for computing gradients is straightforward: first, perform the forward pass, and then, algorithmically perform the backward pass \emph{via} backpropagation (Fig. \ref{fig:geometry}).  
The question then is how to emit continuous derivatives from discrete representations, \aes{especially since the most geometric representations are discrete (\textit{e.g.} bitmaps in 2D and triangle meshes in 3D).}
Three core methods exist for smoothly reasoning about discrete geometry --- 1) smoothing the domain, 2) smoothing the output, and 3) learning continuous representations.

Domain smoothing involves transforming a discrete variable into a continuous one, with the expectation that a downstream algorithm will force its value to a valid integer value.  Examples include topology optimization, which smoothly interpolates voxels between vacancy (0) or occupancy (1), but forces them to 0 or 1 in a downstream algorithm \cite{sigmund2013topology}.  (Neural approaches to such geometry design have also been proposed \cite{sosnovik2019neural, zehnder2021ntopo}.)

Output smoothing occurs when consuming algorithms smooth the domain in order to interpolate the space between discrete regions such as vertices in a mesh.  \aes{C}lassic example\aes{s} include \aes{pixel filters (used in anti-aliasing), and} finite element interpolation, which approximates values at nodes of volumetric meshes across their interior \cite{bhavikatti2005finite}; other more sophisticated methods include multi-resolution kernels \cite{vaxman2010multi} and neural methods such as diffusion models \cite{sharp2020diffusion} and neural subdivisions \cite{liu2020neural}.

\aes{One of t}he fastest growing area in differentiable geometry processing is neural operators, which seek to replace parts of or entire algorithms with domain-agnostic differentiable models that can be learned offline or embedded in a DVC algorithm.  These operators provide useful gradients and can still reason about discrete structures.  One common approach is to embed discrete inputs as continuous tensors that encode the structure of a geometry.
For example, for point clouds, PointNet has seen significant adoption \cite{qi2017pointnet, qi2017pointnet++} (which more generally operates on any unordered set). It has also been modified for implicit point selection \cite{spielberg2021co}, and improved local geometric structures \cite{wang2019dynamic}.  Approaches have also been proposed for deep representations for reasoning about signed distance fields \cite{park2019deepsdf}, manifold graph embeddings \cite{monti2017geometric}, or mesh collections \cite{chang2015shapenet}.  In other cases, neural operators exist that replace entire algorithms. PointFlow provides a multiresolution method for converting meshes to point clouds \cite{yang2019pointflow}.\cite{yang2021geometry} provides a method for mesh deformation using neural fields, and \cite{wang2019deep} proposes a framework for learning generic geometric operators.  

The process of deriving useful gradients in geometry processing shares many of its challenges with physics and rendering; namely, how to continuously reason about discontinuous structure (cf contact in physics and occlusion in rendering), and how to most effectively combine analytical and learned models. In the next section, we describe how gradients are useful for reasoning about the motion and deformation of geometries specified by differentiable representations.

\section{Animation}\label{physics}

\newcommand{\q}{\mathbf{q}}
\newcommand{\param}{\boldsymbol{\theta}}

\begin{figure*}
	\begin{tikzpicture}
		\node[anchor=south west,inner sep=0] at (0,0) { %
		    	\includegraphics[width=\textwidth,page=4,trim=0.0cm 0cm 0.0cm 0.0cm, clip]{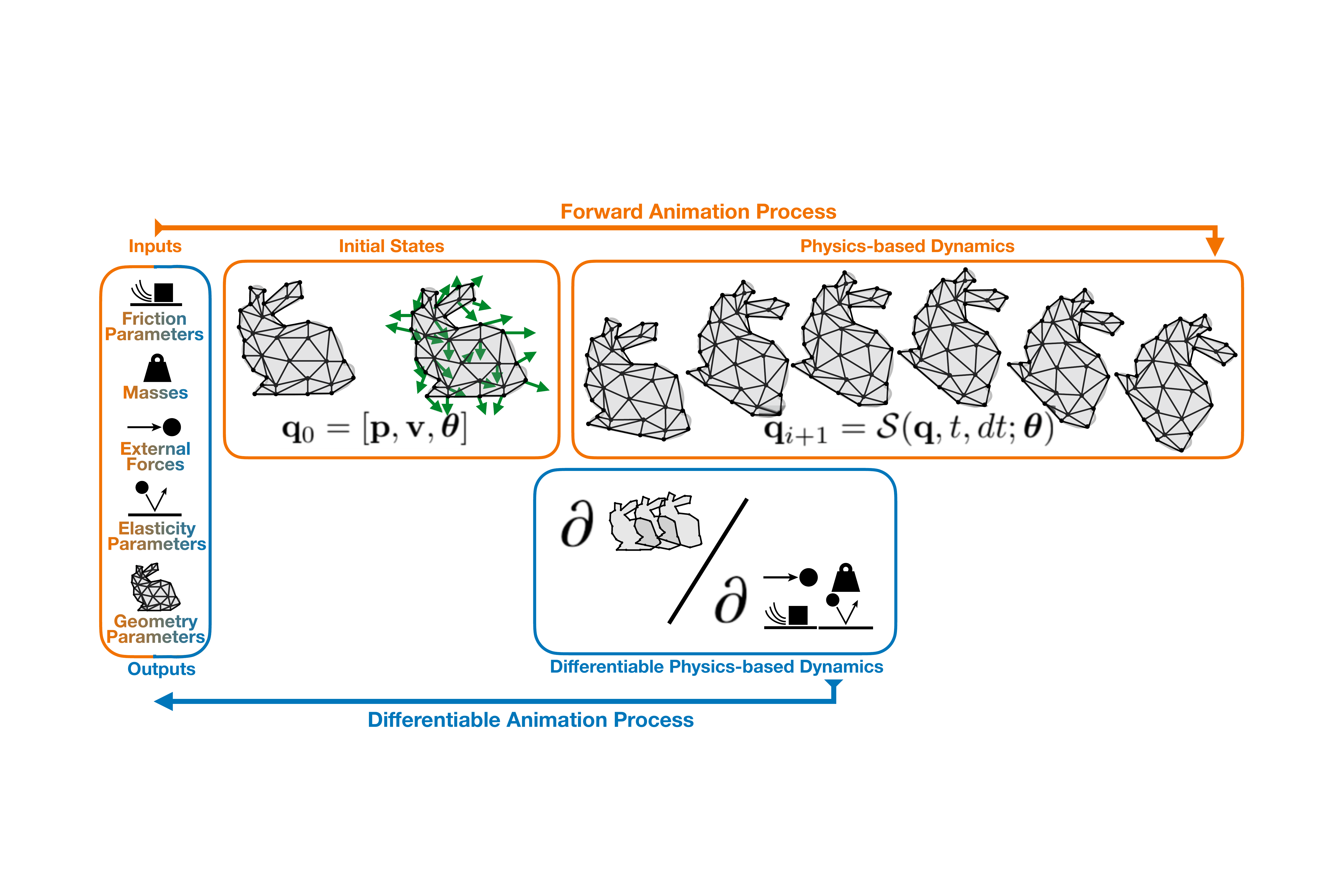}
		}; %
	\end{tikzpicture}
\caption{\label{fig:animation}Forward and differentiable animation. \textbf{Top:} in forward simulation, scene dynamics parameters (far left) are used in a numerical simulation of the time evolution of the dynamics of motion (middle, left to right). \textbf{Bottom:} conversely, in the differentiable animation setting, we aim to compute the gradients of the simulation outputs (i.e., body locations over time, spatial-temporal collision events, etc.) with respect to the dynamics parameters (middle).}
\end{figure*}

\begin{figure*}
\includegraphics[width=\textwidth, page=4, clip, trim=0cm 2cm 6cm 0cm]{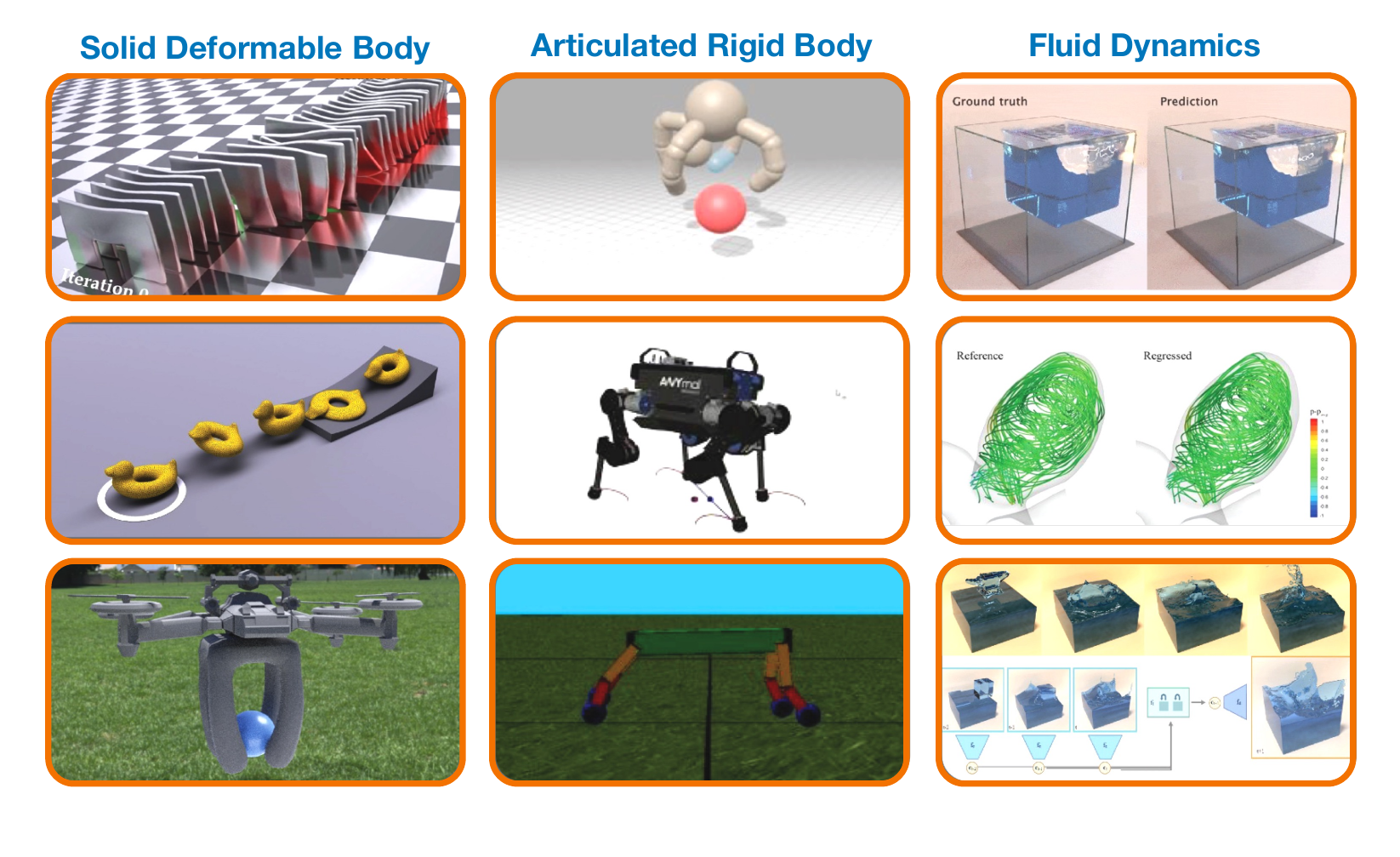}
\caption{Differentiable animation applies to a diversity of problems where system identification and unknown dynamics parameters need be inferred from (i.e., 3D/4D) observations. Application domains include robotics, policy learning, computational manufacturing, and scene understanding.  \aes{\textbf{Papers From Top-To-Bottom; Left Column:} \cite{hu2019chainqueen}, \cite{du2021diffpd}, \cite{qiao2021differentiable}; \textbf{Middle Column:}  \cite{freeman2021brax}, \cite{adrlCT}, \cite{degrave2019differentiable}; \textbf{Right Column:} \cite{sanchez2020learning}, \cite{raissi2020hidden},  \cite{wiewel2019latent}.}}
\label{ref:physicscollage}
\end{figure*}

Physically-based animation, also referred to as physics simulation, provides means of computationally predicting, visualizing, and analyzing the motion and interaction of objects in the world. It is used to design and control industrial machines and robots, predict the behavior of novel materials, and provide high-quality computer-generated imagery for movies and real-time dynamics for digital games.  In this review, we focus primarily on dynamic, time-varying simulators, although static simulation introduces its own interesting challenges for both forward and backward computation \cite{teran2005robust, bacher2021design}.  

Formally, a physics simulation computes the time evolution of a system by computing the change of the system $\dot{\q} = f(\q, t)$ at time $t$, where state variable $\q$ could be compact (in the case of finite rigid degrees of freedom) or infinite in cardinality (in the case of continuum mechanics).  This rate of change is then time-stepped forward \emph{via} time integration.  Since it is intractable to compute the state evolution at every point in time, physical systems are typically time-integrated in discrete steps $dt$.  
Coupling the rate of state change and time integration can be interpreted as a single state transition function $\mathcal{S}$, where $\q_{i+1} = \mathcal{S}(\q, t, dt; \param)$.  Here, $\param$ refers to system parameters, which could include a controller, environmental parameters, design parameters, and so on.

The emerging class of differentiable physics simulators provide a means of both predicting reality and better understanding it.  
While forward simulation allows one to, say, predict a robot's motion, gradients allow one  to efficiently improve it.  
As another application, gradients can allow one to efficiently search for the \aes{assigned} material parameters of a real-world elastomeric object, by searching for \emph{e.g.} stiffness or damping coefficients such that a simulated trajectory matches a real-world measured trajectory.
Compared with corresponding gradient-free methods such as reinforcement learning or evolutionary strategies, gradient-based methods can be orders of magnitude more efficient \cite{hu2019chainqueen, du2021diffpd, dubied2022sim, qiao2021differentiable, de2018end}.

Two prominent strategies exist for constructing differentiable physical simulators --- ``explicit'' methods and ``learning-based'' methods.  Explicit methods derive simulators --- and their gradients --- from first principles, programming a model of the time-evolution of a physical system and algorithmically deriving its gradients.  By contrast, learning-based methods focus on learning a state-transition model from observed simulated or real-world data.  These methods require  less domain knowledge but typically rely on physics-inspired neural network architectures that are catered to the structure of the state evolution \cite{chen2018neural, long2018pde} or matter arrangement \cite{pfaff2020learning, sanchez2020learning}.  In either case, a complete dynamic simulation is constructed by recursively applying the state transition model for the desired duration of the simulation.

Methods for making simulations differentiable are agnostic to the state-transition model, so long as the model itself is differentiable. A common strategy is, given a differentiable state transition model $\mathcal{S}$,  one ``rolls-out'' a system trajectory by iteratively applying $\mathcal{S}$, such that the final state is $\q_T = \mathcal{S}(\mathcal{S}( \ldots \mathcal{S}(\q_0) \ldots ))$~\cite{hu2019chainqueen, de2018end, degrave2019differentiable}.  System state at each intermediate state of the simulation is cached along the way before backpropagating  gradients of a loss $L(\q_T)$ back to the beginning of simulation, with system gradients with respect to states and parameters $\frac{\partial L(\q_T)}{\partial \param}, \frac{\partial L(\q_T)}{\partial \q_{T-1}}$ (See Fig. \ref{fig:animation}). 
This technique has been applied to a large number of domains, including classical (contact-rich) Newtonian mechanics and rigid robotics \cite{de2018end, werling2021fast, freeman2021brax}, elastic continuum mechanics and soft robotics \cite{hu2019chainqueen, sanchez2020learning, du2021diffpd, spielberg2021advanced, qiao2021differentiable}, cloth physics \cite{qiao2020scalable}, and fluid dynamics \cite{wiewel2019latent, raissi2020hidden}, and has been applied for problems such as system identification, robot control learning and optimization, computational design, and dynamical extrapolation.  A summary of some seminal work, and their contributions can be found in Fig \ref{ref:physicscollage}.  

The long time-evolution of physical simulation distinguishes it from geometry and other domains of VC and introduces key challenges.  Vanishing gradients, which also arise when differentiating deep neural networks, can numerically diminish the effect of one variable upon another.  Shortening the gradient horizon and handling the simulation in subsequences is one strategy for overcoming this issue \cite{freeman2021brax}.  Another problem is that of memory consumption, since long simulations require many intermediate states to be stored in memory; this issue can be partially ameliorated by domain-optimal checkpointing strategies \cite{spielberg2021advanced} or decreased precision \cite{hu2021quantaichi}.

In the case of long simulations where one cares about only a few system variables (and memory is precious), ``forward-mode'' differentiation may be employed.  
As states evolve forward, a Jacobian relating gradients of output states to $\q_0$ and $\param$ is maintained and accumulated forward in time, again employing the chain rule.  This occurs ``on the fly''; gradients are computed during the forward simulation pass, never requiring system states to be cached in memory.  Jacobian computation has historically been a workhorse in optimal control (\aes{linear quadratic regulators}, trajectory optimization, \cite{li2004iterative, posa2014direct}), and many robotics toolboxes compute such gradients efficiently \cite{adrlCT, drake}.  Since forward-mode differentiation can only differentiate with respect to a single variable at a time (cf backpropagation and a single loss function), Forward-mode differentiation is ill-suited to high-dimensional input/parameter problems.

Problem structures that are common in animation and control which would be slow or difficult to differentiate have been a major recent focus.  In applications where state evolution relies on an optimization-based controller, or in quasistatic simulation, the loss is the solution of an expensive optimization problem.  In such cases, the implicit function theorem \cite{agrawal2019differentiable} or bespoke optimization ``layers'' \cite{amos2018differentiable} can perform efficient differentiation.  Further, similar to differentiating through discrete structures in geometry, most simulation domains involve contact or other hard boundary conditions; these hard constraints can be repeated many times throughout a simulation.  Choosing a smoothed, relaxed contact model or learning such a model \cite{pfrommer2020contactnets} can alleviate gradient non-smoothness.

\begin{figure*}[t!]
\begin{center}
\includegraphics[width=\textwidth,page=2,trim=0.0cm 0.000cm 0.0cm 0cm, clip]{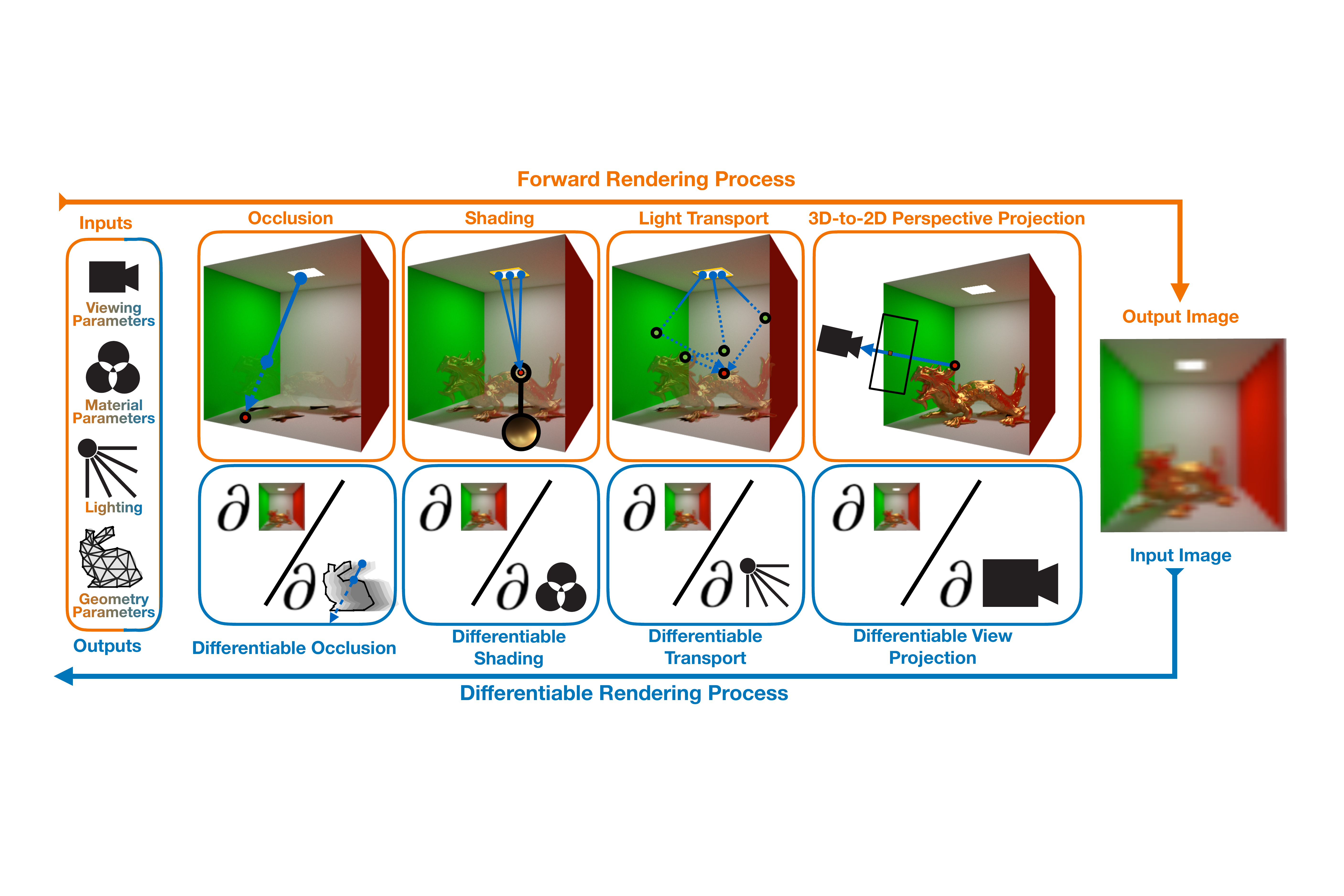}
 
\end{center}
\caption{\label{fig:rendering}Forward and differentiable rendering. \textbf{Top:} when synthesizing an image in forward rendering, scene parameters (far left) are used to synthesize a realistic image (far right), with (middle, left to right) the geometric properties informing occlusion effects, material properties informing reflection and shading, lighting profiles informing complex multi-scattering effects, and viewing parameters defining the imaging sensor's response. \textbf{Bottom:} the differentiable rendering context focuses instead on the numerical estimation of gradients of the synthesized image pixel intensities with respect to the 3D scene parameters, and can be similarly decomposed into the computation of image gradients with respect to: (middle, right to left) changes in the camera sensor's viewing parameters, complex light transport dynamics, localized shading and appearance routines, and occlusion due to potential variations in the scene geometry.}
\end{figure*}
 
 The frontier of differentiable physics is rich with challenges and opportunities.  As dynamics can be particularly chaotic, gradient-based optimization is especially prone to suboptimal local minima here.  Algorithms that combine  exploration (such as those in reinforcement learning or Monte Carlo tree search) with gradient-based optimization remains one exciting horizon \cite{xu2022accelerated}.  More rigorous methods that combine the best of ground-truth and learned models, such as residual neural networks \cite{ajay2019combining, zeng2020tossingbot} or adaptation layers \cite{hwangbo2019learning}, and analyses of the quality and usefulness of learned gradients \cite{ilyas2018closer, metz2021gradients, suh2022differentiable} each provide a path for creating robust models for real-world applications.  
 Finally, we note current physical simulators are usually domain specific --- differentiable multiphysics engines that operate across different time and energy scales are core to modeling the dynamical richness of reality.

\section{Rendering}\label{rendering}

Rendering is the process by geometry and action is seen.  By understanding how (virtual) light reacts with the (virtual) world, the (virtual) world can be observed.  By manipulating light, either through digital displays or physical pigments, the virtual world is then made physical, either through images or animations. 

More formally, given as input a 3D representation of the world \aes{and an assignment of material properties of objects} (\textit{e.g.}, plastic, wood) and the position of light sources and a virtual camera, rendering is the process of converting this 3D representation to a 2D image or 2D images. 
Rendering can encompass a broad range of such processes, spanning those that generate diagrammatic, illustrative, or stylized image representations of the world to those that fully simulate the physical interactions of light and, thus, yield photographic-quality images. The latter is referred to as physically-based rendering (PBR); \aes{for a comprehensive treatment, see \cite{pharr2016physically}.}

\begin{figure}
    \centering
    \begin{subfigure}[t]{0.485\textwidth}
    \centering
    \captionsetup{width=\textwidth}
    \includegraphics[width=\textwidth,page=5,trim=10.0cm 24.5cm 52.5cm 0.0cm,clip]{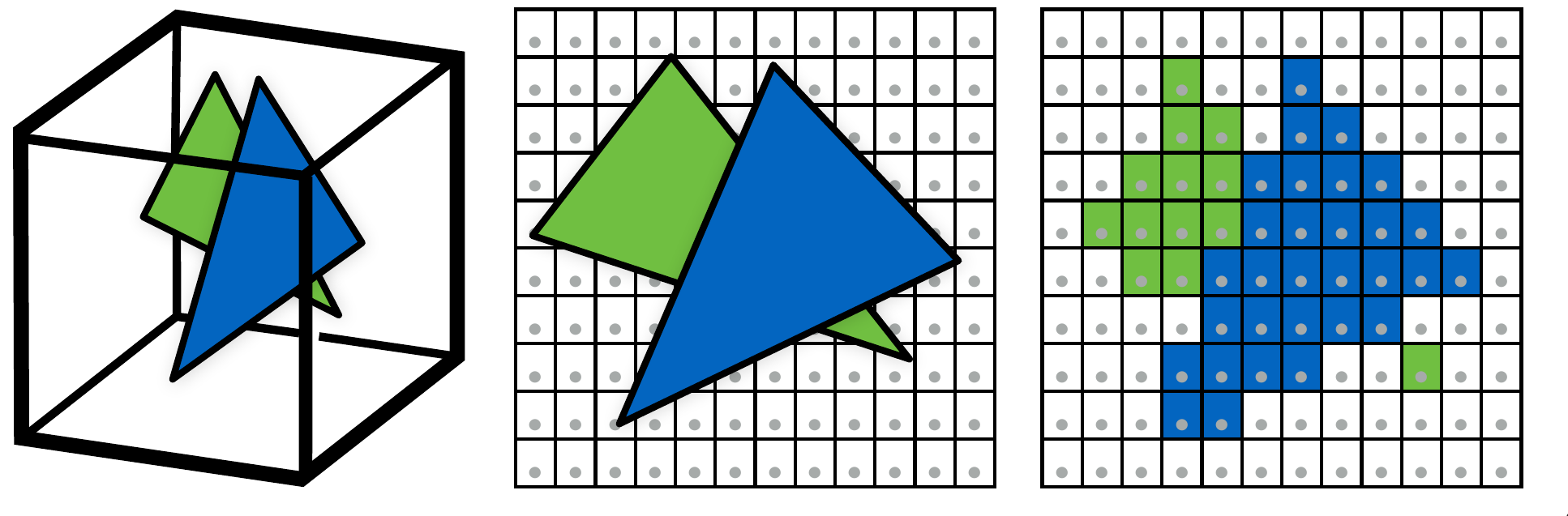}
    \caption{}
    \label{fig:inset1}
    \end{subfigure}
    \begin{subfigure}[t]{0.185\textwidth}
    \centering
    \captionsetup{width=\textwidth}
\includegraphics[width=\textwidth,page=5,trim=42.0cm 24.5cm 43.95cm 0.0cm, clip]{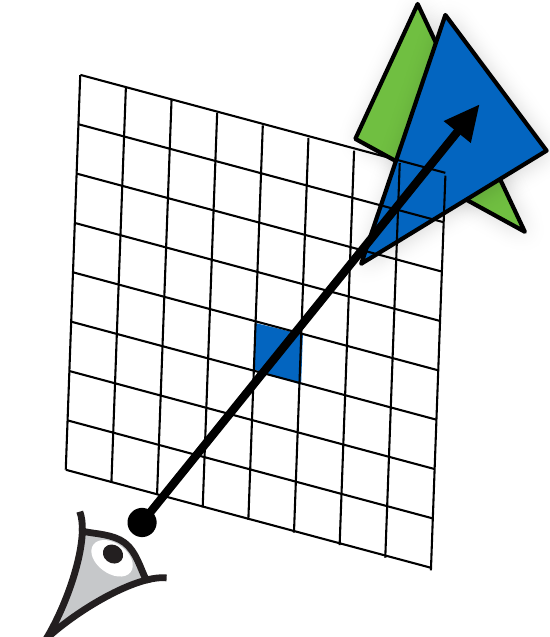}
\caption{}
  \label{fig:inset2}
  \end{subfigure}
  \begin{subfigure}[t]{0.205\textwidth}
  \centering
    \includegraphics[width=\textwidth,page=5,trim=0.0cm 24.5cm 85.5cm 0.0cm, clip]{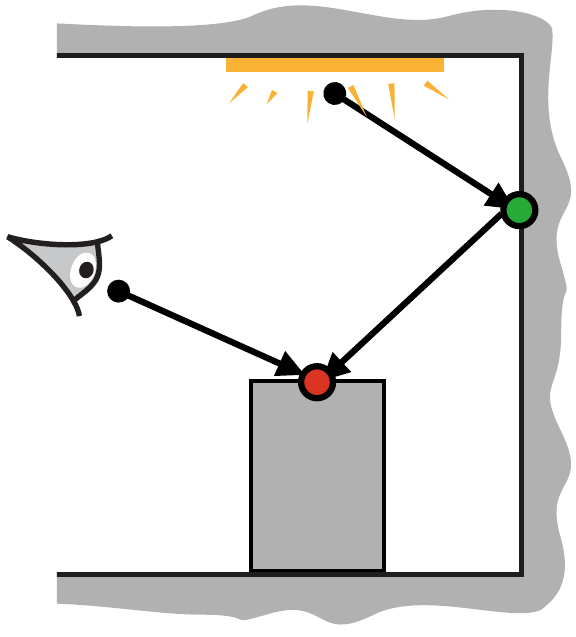}\\
    \captionsetup{width=\textwidth}
    \caption{}
  \label{fig:inset3}
\end{subfigure}
\caption{(a) Projection of triangles in 3D space (left) onto a 2D image plane (center), followed by pixel-wise discretization (right). (b) Ray-tracing images 3D objects by determining where a camera ray would intersect with an object and appropriately colors the image plane. (c) Path tracing is used to realistically model the effects of illumination based on optical principles.}
\end{figure}

VC relies primarily on two algorithmic paradigms when designing rendering algorithms: rasterization and ray-tracing. Rasterization-based frameworks
operate by first performing a sequence of algebraic transformations to warp and project  3D scene elements onto a 2D imaging plane (Fig. \ref{fig:inset1}) before converting the now-2D vector form of the scene elements to a discretized representation on an image buffer. Rendering systems based exclusively on rasterization are favored in high-performance applications, such as video games, as they are more easily parallizable on GPUs. Traded off with this performance is flexibility: implementing PBR numerical methods atop rasterization is markedly harder -- and in some case, impossible -- compared to using ray-tracing. Ray-tracing, as its name implies,  instead obtains a 2D
representation of the 3D world by conceptually tracing rays of light in reverse --- from the (virtual) camera, through each pixel, and into the 3D world --- in order to query which 3D object it first interacts with (Fig.~\ref{fig:inset2} and Fig.~\ref{fig:pipeline}). Ray tracing is better suited to PBR simulations of the underlying physical dynamics of light (called light transport) as it can be used not only to trace rays from the camera, but also from light sources and other objects.  \aes{These simulations help determine which objects are visible to a viewer and  how objects are lit by lights and scene reflections}. Indeed, ray tracing is extended in exactly this manner when simulating the emission, scattering, and absorption of light in a 3D environment.  

The simplest algorithmic realization of this process, path tracing, traces sequences of connected rays, called paths, from the camera and/or light source.  This process is repeated  with stochastic perturbation until an image is synthesized by collecting all light paths that connect a pixel on the camera sensor to an emitter; here, perturbations simulate the various scattering interactions that occur at surfaces in the scene (see Fig.~\ref{fig:inset3} and Fig.~\ref{fig:pipeline}). This process is 
useful not only for studying visible light but any type of electromagnetic radiation\aes{, due to its ray-based nature \cite{dudchik1999microcapillary}}.  

While the forward rendering process takes a fully-defined 3D scene description and generates an image, the inverse rendering process  infers information about the 3D scene given only \aes{image data} as input. Many specialized realizations of inverse imaging have been explored in the foundational literature of computer vision, medical imaging, tomography and even computational astronomy. Most recently differentiable rendering has grown to complement --- and in some cases, subsume --- these more specialized, domain-driven inverse approaches.

Concretely, so-called differentiable rendering approaches solve a variety of sub-problems in inverse inference, namely the numerical estimation of the gradient of a synthesized image with respect to (Figure~\ref{fig:rendering}, left to right) geometrical changes such as occlusion, shading changes due to variability in material reflectance properties, lighting changes due to the multiple scattering of light energy in an environment, and viewpoint changes due to the many possible placements of a (virtual) camera sensor; these events are typically relatively smooth and continuously differentiable except at sparse event points.

The volume of scientific work addressing both end-to-end inverse rendering and its constitutive differentiable rendering sub-problems has grown rapidly in the last handful of years. The evolution of these methods follow similar historical trends as those in the forward rendering problem: development of progressively more refined theoretical models of differentiable rendering \cite{Li:2018:DMC,Loubet2019,Zhang2020,bangaru2021systematically}, paradigm exploration surrounding variants of rasterization \cite{liu2019soft, kato2018neural} and ray-tracing \cite{Li:2018:DMC}, \aes{which include now differentiable ones}, and systems-oriented work to expose these tools to the broader machine learning community \cite{hu2019taichi,hu2019difftaichi,ravi2020pytorch3d,bailey2019differentiable, Jakob2020DrJit}.

The growing ubiquity of neural scene representations, such as Neural Radiance And Density Fields \aes{(commonly shorthanded as NeRF)}\cite{mildenhall2020nerf}, has blurred the lines between differentiable rendering and neural-based approaches to the inverse problem. Here, differentiable rendering formulations custom suited to an underlying (often neural) parameterization of the 3D scene have proven both effective (and necessary!) in tackling the ill-posedness of the inverse scene estimation problem. Image-only supervision, particularly in the low data regime, necessitates strong priors, \textit{i.e.}, surrounding the image formation process.  \aes{For an extensive summary of recent work on neural rendering, please see \cite{tewari2022advances}.}  \aes{Other neural scene representations have also been proposed, including method with higher physical fidelity but not necessarily leading to improved performance in downstream applications \cite{gkioulekas2013inverse, legendre2019deeplight}; when and why certain approaches outperform others remains an open question.}

Open research challenges in differentiable rendering can be roughly categorized along numerical methods and \aes{algorithmic} axes.

The development of more efficient numerical gradient estimators for the various differentiable rendering sub-problems remains an important area of exploration, especially in the context of their applicability to gradient-based optimization. For example, while numerical estimation in the forward rendering context focuses on reducing image synthesis error, the focus of differentiable rendering algorithm development is notably different: variance in the gradient estimates does not necessarily invalidate their utility in gradient-based optimization. In fact, borrowing from many theoretical and practical explorations in the machine learning optimization community \cite{metz2021gradients}, gradient-based optimization methods may actually benefit from variance in minibatched  gradient estimates, as these estimates remain useful when exploring loss landscapes in search of optima that admit lower generalization error.  \aes{Algorithmically, from a computational complexity perspective, new techniques are emerging to address the tradeoff between runtime and the memory footprint in high-dimensional applications \cite{weiss2021differentiable, vicini2021path}.}

\section{Discussion}\label{discussion}

The previous sections \aes{as outlined in Section \ref{pipeline}} presented a vision for a complete DVC pipeline beginning with instantiation of scene geometry, continuing with animation \textit{via} physically-based or -inspired techniques, through to rendering of the simulation to be viewed by humans or computers.  The full differentiability of the pipeline means that gradients of any measurable quantity of the pipeline can be computed with respect to any variable in the system; these gradients allow for hyper-efficient optimization and learning algorithms that can outclass traditional methods and unlock new applications in science and engineering.  For example, such a pipeline could enable effectual  vision systems for autonomous driving, or allow robots to understand the physics of the world from vision and update their controllers in real-time changing environs.  Further, by increasing the efficiency of optimization and learning tasks, learned models are no longer  the application but rather a tool to be used by even more complex algorithms.  If such algorithms could, for example, allow a robot to quickly respond to human interaction, then robots could efficiently plan through predicted interactions, making them more versatile and robust.

This review focused on the three traditional pillars of computer graphics --- geometry, animation, and rendering.
But there remain topics in VC that are less explored and not covered in this survey.  Topics such as texture synthesis \cite{shi2020match}, performance capture \cite{habermann2020deepcap}, and procedural modeling \cite{ritchie2016neurally} are just a few more specific topics which, in recent years, have explored  differentiable methods for improving performance and capability.  Beyond topics which overlap with computer graphics, differentiable methods developed in  VC have the potential to permeate to other aspects of Science, \aes{such as} biology~\cite{alquraishi2021differentiable} \aes{or} molecular dynamics \cite{schoenholz2020jax}.

Despite the promising results presented in this work, DVC as a discipline is in its infancy.  Further algorithmic and systems-level advances will increase the efficiency, usability, applicability, and, ultimately, ubiquity of differentiable methods.  Rather than rely on the ``one simple trick'' of backpropagation, new algorithms and systems will make differentiation more efficient \cite{innes2019differentiable, moses2021reverse}, and lead to the incorporation of other algorithmic methods from Analysis (\cite{bangaru2021systematically}, \cite{bettencourt2019taylor}).  Domain-specific hardware and domain-specific compilers and languages will make it easier to implement new, computationally efficient models, and accelerate research and development times (our most precious resource).  Indeed, such patterns have already begun to emerge with Halide \cite{ragan2013halide, mullapudi2016automatically}, which \aes{supports} chips developed specifically for hardware acceleration on smartphones, \aes{Dr. JIT \cite{Jakob2020DrJit}, a differentiable framework for large computation graphs like those seen in visual computing}, and Taichi \cite{hu2019taichi, hu2019difftaichi, hu2021quantaichi}, a GPU-accelerated differentiable sparse VC language.  
SDKs (such as TensorFlow graphics \cite{bailey2019differentiable} and Pytorch3D~\cite{ravi2020pytorch3d}) that provide quick-start access to the most popular differentiable methods will allow researchers and practitioners to focus on complex problems rather than rote implementation, making DVC models as simple to deploy and integrate as neural networks are today.

DVC is a young field based on an old idea.  In a data-driven world, differentiable methods dare to provide more and ask for less.  They allow humans to insert centuries of scientific knowledge (\emph{i.e.}, human learning) into the DL pipeline, making models more efficient to learn and more easily interpretable.  Throughout the next decade, we expect to see a rich investigation of how to combine neural networks and VC models to extract the best of both approaches, and usher in a new era of machine intelligence that is based on both theory and empiricism.

\aes{
\section{Acknowledgments}
The authors thank Alexander Mordvintsev, Andrea Tagliasacchi, Julien Valentin, Omar Sanseviero, and Sofien Bouaziz for early discussions on content and structure.
}

\aes{
\section{Competing Interests}
The authors declare no competing interests.
}

\bibliography{sn-bibliography}

\end{document}